\documentclass[11pt]{article}

\usepackage[final]{acl}
\usepackage{times}
\usepackage{latexsym}
\usepackage[T1]{fontenc}
\usepackage[utf8]{inputenc}
\usepackage{microtype}
\usepackage{inconsolata}
\usepackage{graphicx}
\usepackage{booktabs}
\usepackage{enumitem}
\usepackage{multirow}
\usepackage{xspace}
\usepackage{url}
\DeclareUrlCommand\filepath{\urlstyle{tt}}
\usepackage[most]{tcolorbox}
\usepackage{fvextra}

\newtcolorbox{promptbox}[1]{colback=green!1!white,
  colframe=green!50!black, fonttitle=\bfseries, title={#1}, left=1mm, right=1mm,
  top=0.5mm, bottom=0.5mm, boxsep=1mm, before skip=4pt, after skip=4pt}
\newtcolorbox{usecasebox}[1]{colback=blue!1!white,
  colframe=blue!45!black, fonttitle=\bfseries, title={#1}, left=1mm, right=1mm,
  top=0.5mm, bottom=0.5mm, boxsep=1mm, before skip=4pt, after skip=4pt}
\newtcolorbox{resultbox}{colback=green!3!white, colframe=green!40!black,
  boxrule=0.5pt, left=1.6mm, right=1.6mm, top=0.8mm, bottom=0.8mm,
  boxsep=0.5mm, before skip=5pt, after skip=5pt, arc=1mm,
  fontupper=\small}
\fvset{fontsize=\scriptsize, breaklines=true, breakanywhere=true,
  breaksymbolleft={}, breaksymbolright={}}

\newcommand{\system}{AgentDebugX\xspace}
\newcommand{\titleicon}{\IfFileExists{figures/agentdebugx_title_icon.png}{\raisebox{-0.18em}{\includegraphics[height=1.35em]{figures/agentdebugx_title_icon.png}}\xspace}{}}

\title{AgentDebugX\titleicon: An Open-Source Toolkit for Failure Observability, Attribution, and Recovery in LLM Agents}

\author{%
  \textbf{Kunlun Zhu}$^{1*}$ \quad \textbf{Xuyan Ye}$^{1*}$ \quad \textbf{Zhiguang Han}$^{1*}$ \quad
  \textbf{Yuchen Zhao}$^{1}$ \quad \textbf{Bingxuan Li}$^{1}$ \quad \textbf{Weijia Zhang}$^{1}$ \\ \quad \textbf{Muxin Tian}$^{2}$ 
 \quad \textbf{Xiangru Tang}$^{3}$ \quad
  \textbf{Pan Lu}$^{4}$ \quad \textbf{James Zou}$^{4}$ \quad
  \textbf{Jiaxuan You}$^{1}$ \quad \textbf{Heng Ji}$^{1}$ \\[2pt]
  $^{1}$University of Illinois Urbana-Champaign \quad
  $^{2}$University of Toronto \quad
  $^{3}$Google \quad
  $^{4}$Stanford University \\
  \texttt{kunlunz2@illinois.edu} \\[1pt]
  {\small $^{*}$Equal contribution.}}

\begin{document}
\maketitle

\begin{abstract}
LLM agent failures are difficult to debug because the step where an error
surfaces is often not the one that caused it. Existing observability tools
replay execution traces but provide little support for identifying the root
cause or translating diagnosis into recovery. We present \textbf{AgentDebugX},
an open-source debugging framework that organizes debugging as a closed loop
of \textbf{Detect}, \textbf{Attribute}, \textbf{Recover}, and \textbf{Rerun}.
At its core, \emph{DeepDebug} performs multi-turn root-cause diagnosis through
global trajectory understanding, structure-guided investigation, and
cross-examination. On the Who\&When benchmark, DeepDebug achieves the best
strict attribution accuracy among the evaluated methods on both tested
open-weight backbones, reaching $28.8\%$ exact agent-and-step accuracy on
\texttt{qwen3.5-9b} versus $21.7\%$ for the strongest single-pass baseline. On
GAIA, DeepDebug repairs $13$ of $73$ failed tasks in a single rerun, compared
with $4$--$6$ for three decoupled self-correction baselines, improving overall
accuracy from $55.8\%$ to $63.6\%$. AgentDebugX exposes this workflow through
a Python library, CLI, web console, and installable agentic skill, and
provides an opt-in Error Hub for sharing scrubbed
failure--diagnosis--repair bundles and reusing them as debugging memory.
\end{abstract}

\begin{table}[t]
\centering
\small
\begingroup
\setlength{\tabcolsep}{2pt}%
\renewcommand{\arraystretch}{1.2}%
\resizebox{\columnwidth}{!}{%
\begin{tabular}{@{}lccccc@{}}
\toprule
 & Port. & Tax- & Attr- & Recov- & Error \\
System & schema & onomy & ibution & ery & hub \\
\midrule
AgentDebug \citep{agentdebug2025} & $\circ$  & $\bullet$ & $\bullet$ & --       & --       \\
MAST \citep{mast2025}             & --       & $\bullet$ & $\circ$   & --       & --       \\
Who\&When \citep{whowhen2025}     & --       & $\circ$   & $\bullet$ & --       & --       \\
AgentDiagnose \citep{agentdiagnose2025} & $\circ$ & $\circ$ & $\circ$ & --      & --       \\
AgentRx \citep{agentrx2026}       & --       & $\bullet$ & $\bullet$ & $\circ$   & --       \\
Langfuse \citep{langfuse2023}     & $\circ$  & --        & --        & --       & --       \\
\textbf{\system (Ours)}                  & $\bullet$ & $\bullet$ & $\bullet$ & $\bullet$ & $\bullet$ \\
\bottomrule
\end{tabular}}%
\endgroup
\caption{Capability coverage versus prior work. $\bullet$ = first-class,
$\circ$ = partial, -- = absent. \textbf{Port.\ schema}: a portable,
framework-agnostic trace format enabling re-analysis, comparison, and
sharing (not a vendor span format). \textbf{Taxonomy}: a labelled
failure-mode vocabulary. \textbf{Attribution}: localizing the responsible
step/agent, not just the crash. \textbf{Recovery}: turning a diagnosis into
a rerun-able fix. \textbf{Error hub}: a shareable cross-team corpus of
diagnosed failures.}
\vspace{-0.2in}
\label{tab:compare}
\end{table}

\section{Introduction}

Large language model (LLM) agents are increasingly deployed in settings that require long-horizon reasoning, external tool use, memory, and coordination across components. They support requests through live APIs, modify
software repositories, operate graphical interfaces, and collaborate through planner--executor and multi-agent workflows~\citep{aghzal2025survey,liu2025comprehensive,2025-metal, yang2025code}. This flexibility also makes their failures difficult to diagnose. As these systems become more capable, however, their failures become substantially harder to debug.\footnote{Project page: \url{https://www.agentdebugx.com}. Code: \url{https://github.com/AgentDebugX/AgentDebugX}. Package: \url{https://pypi.org/project/agentdebugx/} (\texttt{pip install agentdebugx}). Screencast: \url{https://youtu.be/ztni6w0o_l8}. The project is released under the MIT license.}

The central difficulty is that the step where a failure becomes visible is often not the step that caused it. An agent may return an incorrect final answer because of a planning constraint omitted much earlier, a stale memory retrieval, an invalid intermediate assumption, or an incorrect handoff between agents. The resulting symptom may surface only after many apparently reasonable actions, such as a failed tool call, an inconsistent downstream decision, or an unsupported final response. Consequently, replaying an execution trace is rarely sufficient: developers must determine which earlier decision rendered the run unsuccessful, explain why it was decisive, and translate that diagnosis into a correction that can be tested.

Existing tools address only parts of this challenge. General-purpose observability platforms provide detailed execution traces, but largely leave root-cause analysis and repair to the developer. Failure taxonomies and attribution benchmarks formalize common error modes and evaluate whether a method can identify the responsible agent or step, but are typically presented as standalone analyses rather than deployable debugging infrastructure. Self-correction methods can revise unsuccessful behavior, yet correction is considerably more reliable when the location of the underlying error is already known \citep{tyen2024llms}. As summarized in Table~\ref{tab:compare}, current systems therefore leave a gap between observing a failed execution, attributing its root cause, proposing an actionable repair, and verifying that repair through rerunning.

\begin{figure*}[t]
\centering
\includegraphics[width=0.95\textwidth]{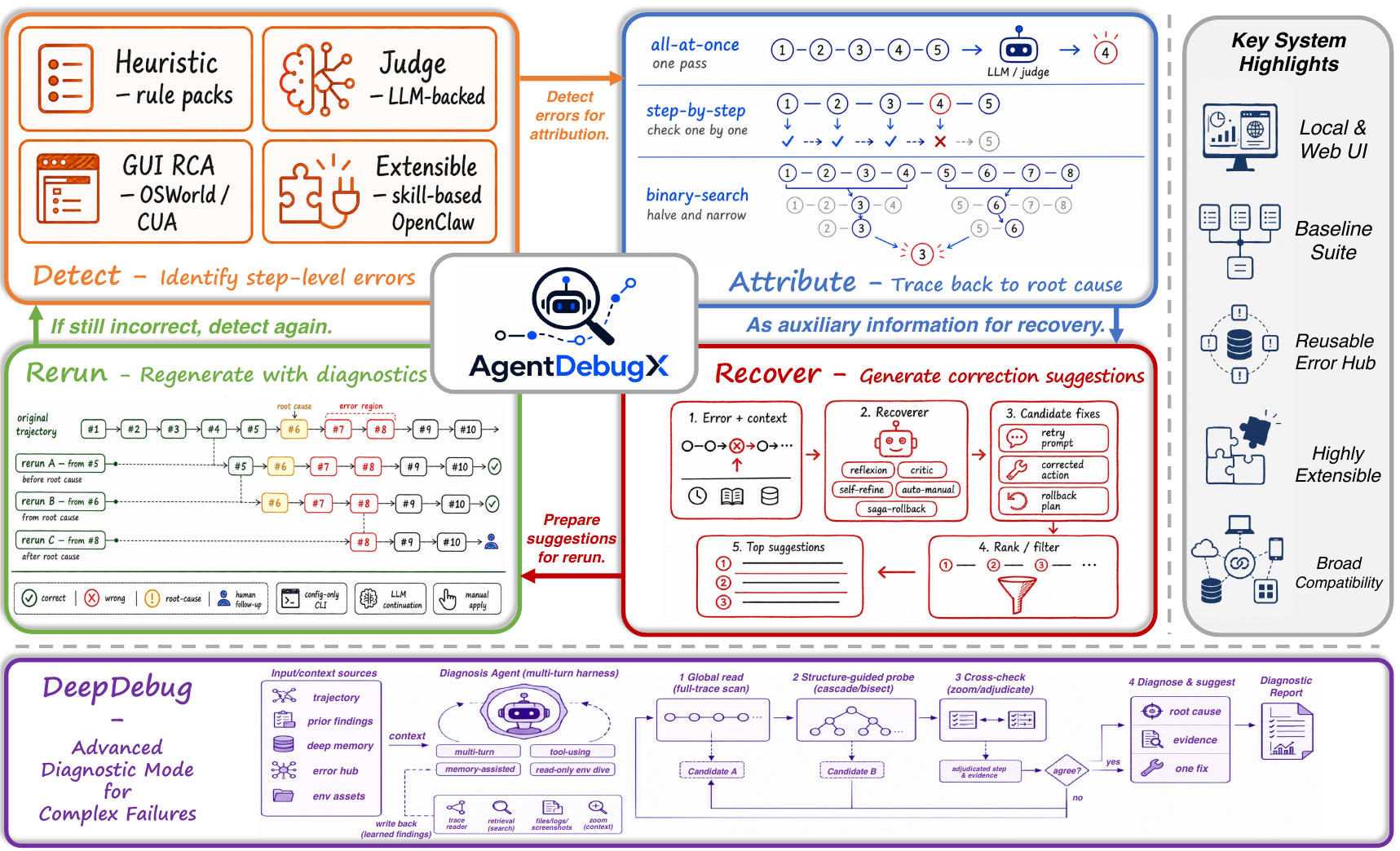}
\caption{\textbf{Overview of AgentDebugX.} \system{} forms a closed debugging loop: \textbf{Detect} identifies step-level errors, \textbf{Attribute} locates their root causes, \textbf{Recover} proposes ranked fixes, and \textbf{Rerun} regenerates the trajectory around the failure point. Hard cases are escalated to \textbf{DeepDebug}, a multi-round diagnosis agent that produces an auditable root-cause report with evidence and a recommended fix. AgentDebugX supports local and web UIs, reusable error sharing, baseline evaluation, and broad agent-framework compatibility.}
\vspace{-0.2in}
\label{fig:architecture}
\end{figure*}

To address the gap, we introduce \system{}, an open-source toolkit that closes this gap by organizing agent debugging as the iterative loop shown in Figure~\ref{fig:architecture}: \textbf{Detect}, \textbf{Attribute}, \textbf{Recover}, and \textbf{Rerun}.
Given either a live execution or an exported log, \system{} first converts framework-specific events into a portable trajectory representation. Detection identifies observable failures and associates them with structured failure modes. Attribution then traces those symptoms backward to the agent and step whose correction would most likely have prevented the failure. Recovery converts the resulting diagnosis into a concrete retry directive, and rerunning applies the approved correction from an appropriate checkpoint while retaining both the original and repaired branches for comparison. When a rerun remains unsuccessful, its new trajectory re-enters the same loop.

At the center of \system{} is \textbf{DeepDebug}, a multi-turn root-cause diagnostic agent designed for failures that cannot be reliably localized through a single trace reading. DeepDebug combines a global trajectory read with a structure-guided probe—tracing handoffs in multi-agent runs or bisecting single-agent traces. It cross-examines conflicting candidates and outputs an auditable report with the responsible agent and step, supporting evidence, an explanation, and one concrete fix. A diagnosis becomes operationally useful only when it can improve the next execution. \system{} therefore connects attribution directly to a policy-gated rerun loop. Its native recovery path uses DeepDebug's localized, evidence-backed correction as the retry directive, while alternative recoverers can reformulate the same diagnosis.

Beyond individual debugging sessions, \system{} provides an opt-in \textbf{Error Hub} for storing scrubbed trajectory--diagnosis--repair bundles. These bundles can serve as incident records, continuous-integration regression fixtures, and reusable debugging memory. Through a shared, framework-independent format, teams can compare diagnoses across methods and versions, retrieve similar historical failures, and accumulate reviewed examples of long-tail failure modes without modifying the original execution evidence.

We evaluate the two capabilities of \system{}: accurately localizing the cause of a failure and converting that diagnosis into a successful repair. On the Who\&When benchmark, DeepDebug achieves the strongest attribution performance among the evaluated methods on both tested open-weight backbones. With \texttt{qwen3.5-9b}, it reaches 28.8\% strict agent-and-exact-step accuracy, compared with 21.7\% for the strongest single-pass baseline. On GAIA, applying DeepDebug's diagnosis in a single rerun repairs 13 of 73 trajectories initially failed by the underlying agent, compared with 4--6 repairs for three decoupled self-correction baselines, increasing overall accuracy from 55.8\% to 63.6\%.

\section{Related Works}
Existing research provides strong components for observing, diagnosing, or correcting agent failures, but rarely connects them into a unified workflow from failure detection to verified recovery.  Observability platforms such as LangSmith, Langfuse, and Phoenix
capture and replay detailed agent traces \citep{agentops2024}, but leave
developers to determine which step was responsible, why, and how to repair the
run. A parallel line formalizes agent failures through taxonomies, attribution
benchmarks, and trajectory diagnostics --- AgentDebug \citep{agentdebug2025},
MAST \citep{mast2025}, Who\&When \citep{whowhen2025}, TRAIL \citep{trail2025},
and AgenTracer \citep{agentracer2025}. These show that even strong models
struggle with root-cause localization, but are presented as standalone
taxonomies, benchmarks, or attribution methods rather than deployable
infrastructure for heterogeneous runtimes. The closest system, AgentDiagnose
\citep{agentdiagnose2025}, provides an open toolkit for scoring trajectories
along interpretable dimensions and curating training data, but does not connect
step-level attribution to verified recovery or a shared incident corpus.
Self-correction methods --- Reflexion \citep{reflexion2023}, Self-Refine
\citep{selfrefine2023}, CRITIC \citep{critic2024}, AutoManual
\citep{automanual2024} --- study how models revise unsuccessful behavior, and
are complementary to our setting: models correct errors far more reliably when
the error's \emph{location} is supplied \citep{tyen2024llms}, which is exactly
what \system{}'s attribution provides.

\section{System Overview}
\label{sec:overview}

Figure~\ref{fig:architecture} illustrates \system{}, a closed-loop debugging framework consisting of four stages: \textbf{Detect}, \textbf{Attribute}, \textbf{Recover}, and \textbf{Rerun}. These stages are coordinated through a portable debugging artifact, while difficult cases are escalated to the DeepDebug agent for diagnosis and recovery.

\subsection{Trace Capture and Representation}
\label{sec:trajectory}

The input to the loop is an \texttt{AgentTrajectory}, a portable record of an
agent execution. The runtime layer converts the events an agent framework emits
--- LLM calls, tool calls and results, memory operations, handoffs, UI actions
--- into an ordered sequence of \texttt{AgentEvent}s. Each event records the
acting agent, module, step index, parent event, inputs, outputs, metadata, and
any error or artifact produced (including screenshots for GUI agents). A
trajectory can be captured from a live runtime through adapters (LangGraph,
CrewAI, the OpenAI Agents SDK, OpenTelemetry, raw ReAct) or reconstructed from
an exported log through offline importers; all mechanisms produce the same
representation, so diagnosis is independent of the original framework. Crucially,
a \emph{diagnosis is layered on top of this record rather than written back into
it}: the same execution can be re-analyzed by any method, compared across
versions, and shared as a regression case without altering the evidence.

\subsection{Closed-Loop Debugging Pipeline}
\label{sec:pipeline}

Given a captured trajectory, \system{} runs the four-stage loop of
Figure~\ref{fig:architecture}. Each stage consumes structured outputs from the
preceding one and produces inspectable artifacts.

\paragraph{Detect.} Deterministic rule packs first target mechanically
verifiable failures --- malformed tool calls, no-progress loops, invalid
outputs, premature success --- with no model call. When rules are insufficient,
an LLM judge reads the goal and a bounded trace window and returns typed
findings (affected event, failure mode, evidence, confidence), expressed in a
shared taxonomy whose seed contains $19$ modes spanning planning, memory, tool
use, verification, and coordination (Appendix~\ref{app:details}). Detection
locates where a failure \emph{manifests}; the detected event is not necessarily
the one responsible, so its findings seed attribution rather than settle it.

\paragraph{Attribute.} The attribution stage traces a symptom to the step that
rendered the task unsuccessful, using a family of strategies that trade cost for
resolution: inexpensive heuristics and single-pass whole-trace reading, then
binary search, per-step inspection, and budgeted ensembles. Each attributor
returns ranked hypotheses with confidence and provenance, not one unqualified
blame assignment, so a deployment dials accuracy against latency and token cost.
Cases that remain ambiguous escalate to DeepDebug (Section~\ref{sec:deepdebug}).

\paragraph{Recover.} Once a responsible step is localized, recovery turns the
finding into a concrete retry proposal grounded in the root-cause step, failure
mode, evidence, and surrounding context. The native path uses DeepDebug's own
correction as the retry directive, requiring no extra model call after
diagnosis; \textbf{Reflexion} \citep{reflexion2023}, \textbf{CRITIC}
\citep{critic2024}, and \textbf{AutoManual} \citep{automanual2024} ship as
alternative strategies and baselines. All are \emph{suggest-only}: because a
repaired action can change the world, application stays behind an explicit
human or policy gate.

\paragraph{Rerun.} \system{} packages the diagnosis, selected checkpoint, and
retry directive into a rerun request; runtime-specific executors consume that
request to produce a new trajectory, and the console additionally supports
model-generated continuation branches for interactive comparison. The branch is
scored against the objective and kept beside the original, preserving both the
failure and the attempted repair; a successful branch is saved as a resolved
case, a failed one re-enters detection. \system{} does not automatically
replay arbitrary external tools from an imported log.

\subsection{DeepDebug: Multi-Turn Root-Cause Agent}
\label{sec:deepdebug}

Single-pass attribution has complementary blind spots: a global read preserves
task context but anchors on the loudest downstream symptom, while a narrow
step-wise scan loses sight of the objective. DeepDebug (lower panel of
Figure~\ref{fig:architecture}) resolves this through a multi-turn, read-only
process over the captured trace, in four stages.

\paragraph{Stage 1 --- global read.} The agent reads the whole trajectory,
reconstructs the objective and history, and names an initial candidate for the
decisive step, keeping the context needed to tell a causal error from a locally
unusual but valid action.

\paragraph{Stage 2 --- structure-guided investigation.} A second pass uses the
strategy matching the trace shape: for a multi-agent run it walks the handoff
cascade upstream from the visible failure to the earliest step that doomed the
run; for a single-agent run it bisects the step range and re-reads the surviving
region, yielding an independent second candidate.

\paragraph{Stage 3 --- cross-examination.} If the two passes agree, the step is
accepted; if they disagree, DeepDebug inspects both candidates side by side with
their context, inputs, outputs, and downstream effects, and selects the stronger
causal explanation --- reducing root-cause selection from a search over the
whole trace to a focused adjudication between two hypotheses.

\paragraph{Stage 4 --- diagnosis and suggestion.} With the step fixed, DeepDebug
emits a structured report: the responsible agent and step, a plain-language
explanation, quoted evidence, and one concrete fix that recovery can use
directly or reformulate. Every inspection is recorded, so the verdict ships with
its audit trail. The agent inspects the trace but never re-executes the run's
tools.

\subsection{Extensible Failure Taxonomy}
\label{sec:induction}

Detection and diagnosis share a structured vocabulary of failure modes. A
fixed taxonomy cannot anticipate every long-tail error, so \system{} proposes
extensions for human review: when the judge meets a recurring failure outside
the seed set it records a novel-mode candidate, and an inducer collects such
residuals, clusters them (label, then lexical or embedding similarity, gated
by a support threshold), proposes one candidate mode per cluster, and
deduplicates against the seed. Proposals \emph{never} overwrite the curated
taxonomy: when several runs show agents waiting on one another indefinitely,
for example, the inducer proposes a new \emph{multi-agent deadlock} mode,
notes its kinship to the existing lost-handoff category, and leaves the
decision to a maintainer.

\subsection{System Surfaces and Integrations}
\label{sec:surfaces}
\label{sec:gui}
\label{sec:hub}
\label{sec:integrations}

\begin{figure}[tb]
\centering
\includegraphics[width=\columnwidth]{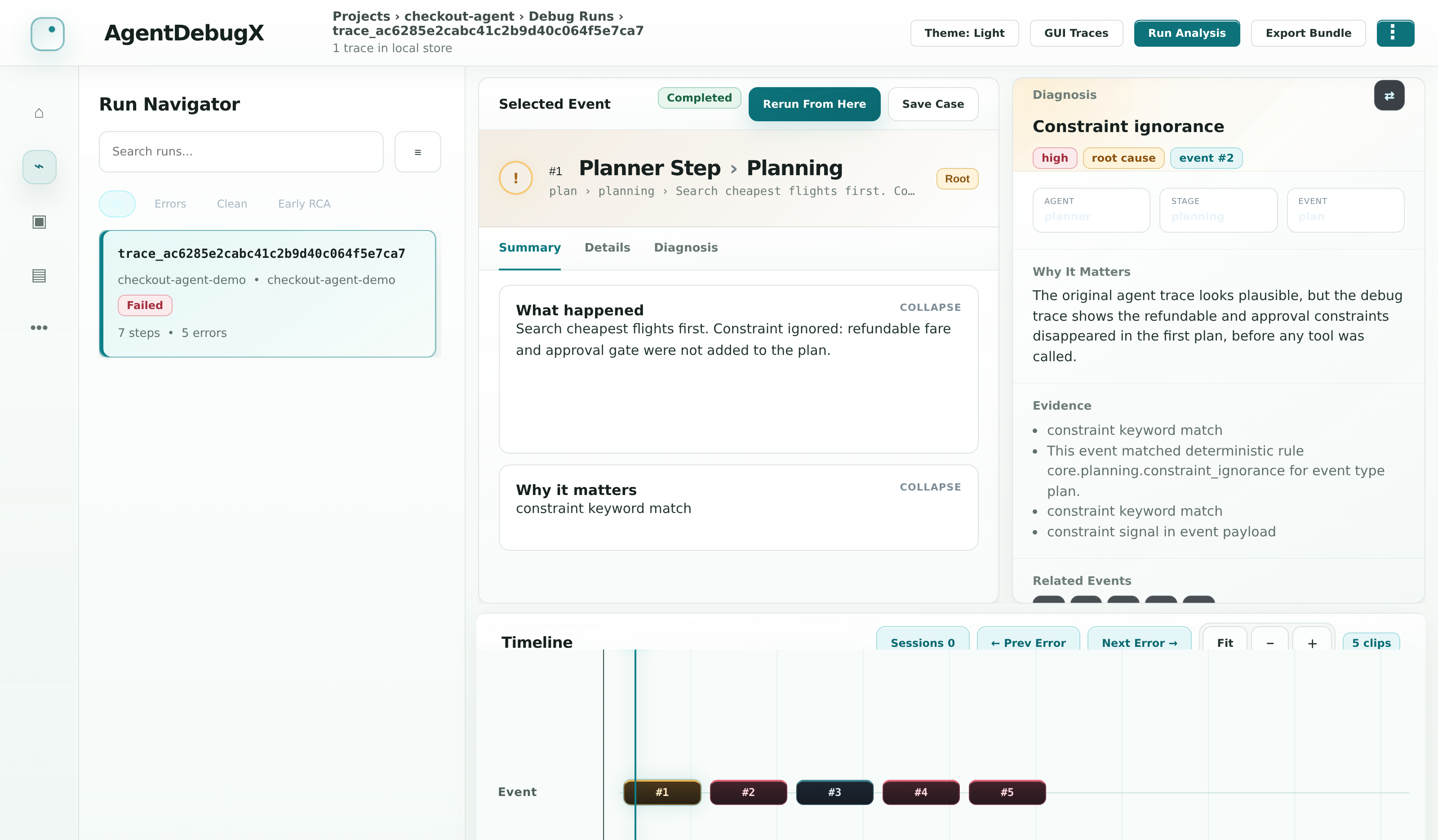}
\caption{The \system{} console on a failed run, as a four-step workflow:
(1)~select a stored trace in the run navigator; (2)~jump from the visible
failure to the attributed responsible event; (3)~read the failure mode,
evidence, and proposed fix in the diagnosis panel; (4)~create a policy-gated
rerun branch and compare it with the original timeline. All artifacts stay in
the local store unless a bundle is explicitly exported.}
\vspace{-0.2in}
\label{fig:console}
\end{figure}

The pipeline is exposed through surfaces that share the same trajectory,
finding, report, and recovery types, so a case created in one can be continued
in another. The \textbf{interactive console} (Figure~\ref{fig:console}) launches with
\texttt{agentdebug serve} over the local store the library writes; as a
no-build single-page app shipped in the wheel, it takes a developer from
\texttt{pip install} to a live debugger. A typical session follows the four
numbered steps of Figure~\ref{fig:console}: select a failed run, jump from
the visible failure to the attributed responsible event, review evidence and
the proposed fix, then fork a \emph{rerun-from-step} branch that is scored
against the original. The same views serve
computer-use agents: an OSWorld importer normalizes screenshot-and-action steps
and a GUI root-cause mode reasons over the visual channel, while a Python-style
traceback or JSON report serves CI when no browser is wanted.

The \textbf{Error Hub} turns incidents into shared knowledge. A trajectory, its
report, and artifacts are packed into a bundle whose scrubber, by default,
strips event inputs wholesale --- the fields carrying prompts and tool
arguments --- and applies known-pattern credential and PII redaction to every
remaining string; because pattern redaction cannot guarantee removal of
arbitrary sensitive content, sharing stays opt-in and bundles should be
reviewed before publication. Bundles are written to a local directory, a
private Git remote, or a public dataset, doing double duty as a CI fixture
and an entry in a cross-team corpus. That corpus is also \system{}'s
long-term \textbf{memory}: DeepDebug can retrieve similar past cases to seed
its hypotheses and writes each new case back; through the shared bundle
format, accepted entries and reviewed taxonomy additions become available to
future diagnosis (this effect is not yet evaluated). Finally, DeepDebug is
packaged as an installable \textbf{agentic skill} and CLI, so tool-using agents
(Claude Code, OpenClaw, and Hermes, each installing the skill) can normalize their
own failed run, diagnose it, and feed the fix back into their next attempt ---
one CLI contract through which supported host agents can diagnose their own runs and each other's.

\section{Evaluation}
\label{sec:eval}

We evaluate the two things a deployed debugger must do, in pipeline order:
localize the cause accurately, then turn that diagnosis into a \emph{fix}.

\paragraph{Setup.} Unless noted, the debugged policy is \texttt{qwen3.5-9b}
and all diagnosis (judge and DeepDebug) runs on \texttt{gemini-2.5-flash},
at temperature $0$ with thinking disabled.
Diagnostic memory and the Error Hub stay \emph{empty} throughout. Per-task
outputs, configs, and a script regenerating Table~\ref{tab:gaia} ship with
the code; full protocol in Appendix~\ref{app:evaluation}.

\paragraph{Failure attribution.}
We evaluate on all $184$ Who\&When traces \citep{whowhen2025}. Each trace is
labelled with \emph{who} erred --- the \emph{responsible agent}, whose
decision doomed the run --- and \emph{when} --- the exact \emph{mistake
step}. Following the benchmark's
reference-answer protocol, every method receives the task's reference answer
but neither gold label. With \texttt{qwen3.5-9b} (Table~\ref{tab:attr}),
DeepDebug's two-reading adjudication outperforms the evaluated
single-strategy localizers on all five reported metrics --- responsible agent
($56.0$ vs $47.8$), exact/near ($\pm1$) step ($28.8$/$44.0$ vs $22.3$/$38.6$),
and strict agent-and-step ($28.8$/$32.1$ vs $21.7$/$23.9$). The joint gain
carries to \texttt{qwen3.6-27b} (strict $38.0$ vs $36.4$). The benefit is
model-dependent: on the hosted backbones of Appendix~\ref{app:evaluation} a
single global reading is already stronger and adjudication does not improve
it, motivating per-model routing rather than invoking the multi-call method
unconditionally. Swapping the structure-aware reading for a second global searcher costs
$4.8$ strict points on \texttt{gpt-5.4-mini} (ablation,
Appendix~\ref{app:evaluation}).

\begin{figure}[tb]
\centering
\includegraphics[width=0.9\columnwidth]{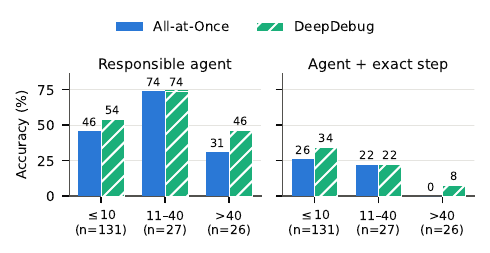}
\caption{Who\&When accuracy by trace length ($n{=}184$;
\texttt{qwen3.5-9b}). Left: responsible-agent accuracy (\emph{who}).
Right: strict joint accuracy requiring both the responsible agent and the
exact mistake step (\emph{who}+\emph{when}) to be correct.}
\label{fig:bylength}
\vspace{-3mm}
\end{figure}

\paragraph{Where the multi-turn agent pays off.}
Figure~\ref{fig:bylength} breaks the same run down by trace length: the
margin concentrates on traces longer than $40$ events, the regime the
structure-guided turn and cross-examination target. Only $26$ traces fall
in this range, so we treat it as descriptive evidence.

\begin{table}[t]
\centering
\footnotesize
\setlength{\tabcolsep}{3.4pt}
\begin{tabular}{@{}l ccccc@{}}
\toprule
Method & Agent & Step & Step & A+S & A+S \\
       &       & (exact) & ($\pm1$) & (exact) & ($\pm1$) \\
\midrule
Rule heuristic & 14.1 & 1.6 & 6.5 & 1.6 & 2.7 \\
All-at-Once  & 47.8 & 22.3 & 35.3 & 21.7 & 23.9 \\
Step-by-Step & 41.8 & 18.5 & 38.0 & 17.9 & 19.6 \\
Binary-Search& 41.8 & 17.4 & 38.6 & 17.4 & 20.1 \\
\textbf{DeepDebug} & \textbf{56.0} & \textbf{28.8} & \textbf{44.0} & \textbf{28.8} & \textbf{32.1} \\
\bottomrule
\end{tabular}
\caption{Failure attribution on the full Who\&When benchmark ($n{=}184$;
\texttt{qwen3.5-9b}). ``Agent'' measures responsible-agent accuracy
(\emph{who}); ``Step'' measures mistake-step localization (\emph{when})
under exact and $\pm1$ criteria; ``A+S'' requires both agent and step to
be correct.}
\label{tab:attr}
\vspace{-3mm}
\end{table}

\paragraph{End-to-end recovery on GAIA.}
Localization is only useful if it \emph{changes an outcome}, so we close the
loop. A vanilla \texttt{qwen3.5-9b} Open-Deep-Research agent solves $55.8\%$ of
GAIA validation and fails $73$ tasks; we diagnose each failure and rerun it
once. \system{}'s \textbf{native path} phrases DeepDebug's own localized fix
directly into the retry; Reflexion, CRITIC, and AutoManual run as
\emph{decoupled} baselines: same failed-task context plus a generic
judge-produced failure summary, but not DeepDebug's localization, evidence, or
authored fix. Under this shared subset, DeepDebug's recovery produces more
than double the one-rerun repairs of the decoupled baselines
(Table~\ref{tab:gaia}), with the largest gain on Level-2 multi-hop tasks
($48.8\rightarrow61.6$) --- evidence its localized, evidence-backed fix is
useful, though the experiment evaluates the full recipe rather than isolating
attribution alone.

\begin{table}[t]
\centering
\small
\setlength{\tabcolsep}{3.4pt}
\begin{tabular}{@{}lccccc@{}}
\toprule
Method & Rec. & L1 & L2 & L3 & All \\
\midrule
Vanilla \texttt{qwen3.5-9b} & --- & 77.4 & 48.8 & 34.6 & 55.8 \\
\midrule
\quad{+}\,CRITIC & 4/73 & \textbf{81.1} & 51.2 & 34.6 & 58.2 \\
\quad{+}\,AutoManual & 5/73 & 79.2 & 53.5 & 34.6 & 58.8 \\
\quad{+}\,Reflexion & 6/73 & 77.4 & 55.8 & 34.6 & 59.4 \\
\quad{+}\,\textbf{DeepDebug (ours)} & \textbf{13/73} & \textbf{81.1} & \textbf{61.6} & 34.6 & \textbf{63.6} \\
\bottomrule
\end{tabular}
\caption{End-to-end recovery on GAIA validation ($165$ tasks; official
scorer). ``Rec.'' denotes the number of repaired cases among the $73$
vanilla-failed trajectories. Accuracy (\%) is reported by difficulty
level and overall after a single rerun.}
\label{tab:gaia}
\vspace{-3mm}
\end{table}

\paragraph{Cost-aware selection.} Every backend declares its inference cost
(free rules; one call for All-at-Once and the judge; $O(\log N)$/$O(N)$
searches; $\sim\!5$ calls for DeepDebug), letting deployments trade model
calls for localization accuracy. The escalation is cheaper than the call count
suggests: on a $25$-trace stratified Who\&When sample, one whole-trace pass
averages $8.1$K tokens against DeepDebug's $12.8$K ($1.6\times$; its later
turns read focused windows), so escalating only ambiguous cases keeps
expected cost near a single pass; every attributor returns ranked hypotheses with
provenance rather than one authoritative blame claim.

\begin{resultbox}
\textbf{Takeaway.} Across both experiments, DeepDebug's multi-turn diagnosis
delivers state-of-the-art attribution on Who\&When ($28.8\%$ strict vs
$21.7\%$ for the best single strategy) and turns those diagnoses into
$2$--$3\times$ more repaired GAIA failures than decoupled self-correction
($+7.9$ points), at only $1.6\times$ the tokens of a single pass.
\end{resultbox}

\section{Use Cases and Applications}
AgentDebugX supports a range of debugging workflows. A typical session begins
by capturing or importing a failed execution. DeepDebug then localizes the
responsible step, explains the root cause, and proposes a repair, which can be
reviewed before a policy-controlled rerun. The resulting trajectory is retained
alongside the original and can optionally be stored in the Error Hub as
reusable debugging memory. This workflow supports interactive debugging, CI
regression testing, incident response, and self-debugging agents.

\section{Conclusion}
In this work, we presented \system{}, a closed-loop debugging framework that connects failure detection, root-cause attribution, recovery, and rerun for LLM agents. Its core method, DeepDebug, demonstrates that improving fault attribution can translate into measurable gains in downstream recovery. We hope AgentDebugX provides a practical foundation for debugging increasingly capable LLM agents and facilitates future research on reliable agent development.

\section*{Ethical Considerations}

\system collects potentially sensitive agent traces, including prompts, tool
arguments, user data, files, screenshots, and outputs. The default deployment is
local-first, and shared corpus upload is opt-in. Any production deployment
should configure redaction, retention, access control, and audit logging before
collecting user data. Diagnostic labels and recovery suggestions can be wrong,
so the UI and API should present confidence and evidence rather than framing
attribution as ground truth. For high-impact domains, deployments should require human or policy approval
rather than automatic recovery.

\section*{Broader Impact}

\system{} can help make agent reliability an inspectable and measurable
engineering practice. Today, failures are often patched privately and then
forgotten; systematic attribution and recovery evidence can instead inform
audit trails, deployment gates, regression tests, and decisions about when an
agent should not be trusted. This matters as agents enter software development,
science, education, accessibility, and public-facing services, where silent or
repeated errors can propagate beyond a single run. An open-source toolkit and
common trace representation also lower the cost for researchers and smaller
organizations to study robustness and compare systems on shared evidence.

The \textbf{Error Hub} extends this impact from individual deployments to a
community: one team's failure can become another team's regression case and,
at scale, contribute to realistic benchmarks, evolving taxonomies, and better
diagnostic or recovery methods. Together, AgentDebugX and the Hub create a
path from isolated incidents to organizational learning and eventually shared
robustness standards. Realizing this vision requires consent, provenance,
effective redaction, moderation, takedown processes, and audits of whose
systems and failures remain underrepresented. With these safeguards,
AgentDebugX could make agent robustness more cumulative, collaborative, and
broadly accessible rather than a proprietary advantage.

\bibliography{custom}

\clearpage
\appendix

\section{System and Prompt Details}
\label{app:details}
\label{app:prompts}

\paragraph{Trace schema.} Each run is an \texttt{AgentTrajectory} of
\texttt{AgentEvent}s (type, agent, module, step index, parent, timestamp,
inputs/outputs, error, duration, metadata, artifacts). Artifacts can hold text,
images, audio, UI state, files, or environment snapshots, and the same events
project to OpenTelemetry GenAI spans (\texttt{invoke\_agent}, \texttt{chat},
\texttt{execute\_tool}, \texttt{handoff}). The diagnosis layers are driven by
compact, JSON-constrained prompts; we reproduce the most load-bearing ones
below (the full prompt library ships in the repository), followed by a
complete diagnostic report produced during the GAIA experiment.

\begin{promptbox}{LLM Judge --- System Prompt (abridged)}
\begin{Verbatim}
You are AgentDebugX-Judge. Given the goal, the
ALLOWED failure-mode codes, and one run's events,
label each failed step with ONE allowed code. Be
conservative; return an empty list if no failure.
Output ONLY JSON, no prose/fences; cap findings at
{max_findings}; evidence < 120 chars; never emit
newlines inside strings; close the JSON completely.
Schema: {"findings":[{"event_id","step_index",
"agent_name","failure_mode_id","confidence",
"evidence":[...]}], "summary":"..."}
\end{Verbatim}
\end{promptbox}

\begin{promptbox}{Whole-Trace Localizer --- System Prompt (abridged)}
\begin{Verbatim}
You are an AI assistant tasked with analyzing
agent conversation history when solving a real
world problem.
Respond ONLY with a JSON object matching this
schema (no prose, no markdown):
{ "span_id": "<event_id or null>",
  "step_index": <int or null>,
  "agent_name": "<agent_name or null>",
  "confidence": <float 0..1>,
  "rationale": "<one or two sentences>",
  "evidence": ["<short quoted evidence>", ...] }
If the trajectory does not appear to have failed,
return all fields as null.
\end{Verbatim}
\end{promptbox}

\begin{promptbox}{Cross-Examination (disagreement arbitration)}
\begin{Verbatim}
System: A run FAILED. Two candidate error steps
are shown with their context. Which ONE is the
decisive critical error step (the one whose
correction would most likely have averted the
failure)?
Respond ONLY with JSON: {"step": <int, one of
the two>}

User: Goal: <task goal>
Correct answer: <reference, when available>
CANDIDATE A = step <i>: <+-1-step context>
CANDIDATE B = step <j>: <+-1-step context>
Which is the critical error step, <i> or <j>?
\end{Verbatim}
\end{promptbox}

\begin{promptbox}{DeepDebug Refine --- System Prompt}
\begin{Verbatim}
You are AgentDebugX-DeepDebug writing the final,
human-readable diagnosis. The decisive ROOT-CAUSE
step has ALREADY been localized for you -- do NOT
second-guess or move it. Using the step and its
surrounding context, write: a one-or-two sentence
diagnosis of WHY that step caused the failure;
the concrete evidence (short quoted snippets from
the shown steps); and ONE concrete, actionable
fix.
Respond ONLY with JSON: {"summary": "<1-2
sentences>", "evidence": ["<short quoted
snippet>", ...], "suggestion": "<one fix>"}
\end{Verbatim}
\end{promptbox}

\paragraph{Example diagnostic report.} The abridged report below is DeepDebug's
actual output on a GAIA validation task the vanilla agent failed (a
statistics question over 1{,}002 papers with average $p$-value $0.04$); the
agent had assumed $p$-values were uniformly distributed to estimate the
number of incorrect papers. Fed verbatim as the retry directive, this
diagnosis led the rerun to the correct answer ($41$), one of the $13/73$
recoveries in Table~\ref{tab:gaia}.

\begin{usecasebox}{DeepDebug report (GAIA task 04a04a9b, abridged)}
\begin{Verbatim}
{ "trace_id": "04a04a9b-...",
  "root_cause_agent": "print",
  "root_cause_step_index": 9,
  "findings": [{
    "failure_mode": "attribution.root_cause",
    "step_index": 9, "confidence": 0.6,
    "evidence": [
      "# Assuming uniform distribution
         [0, 2*average_p_value]",
      "proportion_incorrect = (2*average_p_value
         - threshold) / (2*average_p_value)"],
    "suggestion": "Calculate the number of
      incorrect papers using ... the direct
      definition of the p-value as the Type I
      error rate under the null hypothesis:
      1002 * 0.04 = 40.08, rounding up to 41."
  }],
  "summary": "The model incorrectly assumed a
    uniform distribution of p-values over
    [0, 0.08] to calculate the proportion of
    incorrect papers -- a mathematically
    arbitrary assumption not supported by the
    problem statement.",
  "metadata": { "analyzer": "DeepDebugAnalyzer",
    "backend": "aao_moe", "memory_used": false }}
\end{Verbatim}
\end{usecasebox}

\section{Deployment Requirements}
\label{app:requirements}

The design follows five requirements from production agent operations:
\textbf{low-friction} capture (a context manager, callbacks, or importers);
a \textbf{portable} representation with an OpenTelemetry GenAI export path
\citep{otelgenai}; \textbf{typed} diagnoses (root cause, evidence,
confidence, fix); \textbf{local-first} storage with explicit scrubbing before
sharing; and \textbf{cost-aware} analysis, where deterministic triage is free
and LLM depth is opt-in.

\section{Evaluation Protocol}
\label{app:evaluation}

\paragraph{Benchmarks.} \emph{Who\&When} \citep{whowhen2025} is used in full
($n{=}184$: $126$ algorithm-generated, $58$ hand-crafted), each trace annotated
with the gold responsible agent and mistake step. \emph{GAIA} \citep{gaia2023} validation ($165$
tasks across three difficulty levels) is used for end-to-end recovery, scored by
the official question scorer.

\paragraph{Attribution protocol.} Following the Who\&When ``with ground truth''
setting, every method receives the task's reference answer as context; none
sees the gold agent or step. Decoding is at temperature $0$ with model-side
thinking disabled; completions are capped at $4{,}096$ tokens ($512$ for
arbitration). Agent names are normalized before comparison, and step match is
exact index equality. We report responsible-agent accuracy, exact and $\pm1$
step-localization accuracy, and the joint agent-and-step (A+S) metrics. Backbones
span open-weight (\texttt{qwen3.5-9b}, \texttt{qwen3.6-27b}) and hosted
(\texttt{gpt-5.4-mini}, \texttt{gemini-3.5-flash}) models, all served through an
OpenAI-compatible endpoint.

\paragraph{GAIA recovery protocol.} A vanilla Open-Deep-Research agent
(\texttt{qwen3.5-9b} policy, Gemini-2.5-flash search) is run once over all $165$
validation tasks, yielding a $73$-task failure subset. Each failed trajectory is
converted to an \texttt{AgentTrajectory}, diagnosed by the LLM judge and, for the
native path, DeepDebug. Recovery then reruns \emph{only} the $73$ failed tasks
once, under one of four strategies: \system{}'s native DeepDebug directive, or
the decoupled Reflexion / CRITIC / AutoManual baselines (which see a generic
judge summary, not DeepDebug's localized diagnosis). To keep the comparison
clean, the diagnostic memory and Error Hub are empty during evaluation; both are
opt-in for real deployments. Reruns use temperature $0$, a fixed step budget, and
the same search and tool stack as the baseline. ``Rec.''\ counts repaired tasks
over the $73$; ``All''\ projects corrected tasks over the full $165$.

\paragraph{Localizer design ablation.} DeepDebug's turn design was chosen by
measurement on a stratified $42$-trace Who\&When sample: replacing the
structure-guided turn with a second global search drops strict accuracy from
$0.310$ to $0.262$ on \texttt{gpt-5.4-mini}, while the shipped pairing lifts
agent accuracy to $0.524$ vs $0.429$ for one reading; on
\texttt{gemini-3.5-flash} a single reading is already strongest ---
adjudication pays on the evaluated open-weight backbones but not on the
hosted ones --- a model-dependent effect.

\section{Implementation}
\label{app:impl}
\label{sec:impl}

\system{} is a dependency-light, MIT-licensed Python package: \texttt{pip
install agentdebugx}, imported as \texttt{agentdebug}. The public API centers on
\texttt{AgentDebug}, \texttt{TraceSession}, \texttt{AgentTrajectory},
\texttt{FailureFinding}, and \texttt{DiagnosticReport}; adoption is one
context manager:

\begin{quote}
\small
\begin{verbatim}
from agentdebug import AgentDebug, EventType
dbg = AgentDebug()
with dbg.trace(goal="Book flight",
               framework="my-agent") as t:
    t.record(EventType.PLAN, agent_name=
      "planner", output="Search fares")
    t.record(EventType.TOOL_RESULT,
      agent_name="browser", step_index=3,
      error="Checkout timeout")
    report = t.analyze()
\end{verbatim}
\end{quote}

Traces persist to append-only JSONL or SQLite. Three capture surfaces emit
the same schema: runtime adapters (raw ReAct, LangChain/LangGraph, CrewAI,
OpenAI Agents SDK, OpenTelemetry GenAI), offline importers (message lists,
conversations, event lists, WebShop pages, OpenAI Agents spans, CrewAI
events, OpenClaw sessions), and host integrations (a Claude Code skill and a
CLI skill contract for tool-using agents), so detection, attribution, and
recovery are source-independent. The CLI exposes the workflow as
\texttt{ingest / diagnose / inspect / act}, with \texttt{serve} for the
console.

\section{Limitations}
\label{app:limitations}

Our evaluation covers automatic attribution and recovery, not developer
debugging time or console usability. Who\&When uses the benchmark's
reference-answer protocol, and DeepDebug's gains vary by model, so its extra
calls are not uniformly beneficial. The GAIA experiment evaluates one
policy model on a fixed failed subset and compares complete retry recipes; it
neither isolates attribution's effect alone nor is a blind single-shot score.
Error Hub retrieval and taxonomy induction are implemented but not yet
evaluated, and induction requires human acceptance. The scrubber strips event inputs and redacts known
credential/PII patterns but not arbitrary content; recovery execution stays
gated on human approval.

\end{document}